\documentclass{article}

\usepackage{arxiv}

\usepackage[utf8]{inputenc} % allow utf-8 input
\usepackage[T1]{fontenc}    % use 8-bit T1 fonts
\usepackage{hyperref}       % hyperlinks
\usepackage{url}            % simple URL typesetting
\usepackage{booktabs}       % professional-quality tables
\usepackage{amsfonts}       % blackboard math symbols
\usepackage{nicefrac}       % compact symbols for 1/2, etc.
\usepackage{microtype}      % microtypography
\usepackage{lipsum}
\usepackage{float}% If comment this, figure moves to Page 2
\usepackage{graphicx}
\usepackage[export]{adjustbox} %for large figures
\usepackage{epsfig}
\DeclareGraphicsExtensions{.pdf,.jpeg,.png}
\usepackage{multirow}
\usepackage{makecell} % for more vertical space in cells
\usepackage{booktabs}
\usepackage{caption}
\usepackage{amsmath}
\usepackage{subfig}
\usepackage{csvsimple}
\usepackage{longtable}
\usepackage{multicol}

\newsavebox\ltmcbox

\newcounter{entryno}
\def\tabline{Test & \the\value{entryno} & Description\addtocounter{entryno}{1}\\}

\title{Machine: The New Art Connoisseur}

\author{
  Yucheng Zhu\\
  McCormick School of Engineering\\
  Northwestern University\\
  Evanston, IL, USA\\
  \texttt{zhu@u.northwestern.edu}\\
   \And
 Yanrong Ji\\
  Feinberg School of Medicine\\
  Northwestern University\\
  Chicago, IL, USA\\
  \texttt{yanrong.ji@northwestern.edu}\\
   \And
 Yueying Zhang\\
  McCormick School of Engineering\\
  Northwestern University\\
  Evanston, IL, USA \\
  \texttt{yueyingzhang2019@u.northwestern.edu}\\
   \And
 Linxin Xu\\
  McCormick School of Engineering\\
  Northwestern University\\
  Evanston, IL, USA\\
  \texttt{linxinxu2019@u.northwestern.edu}\\
 \And
 Aven Le Zhou\\
  New York University, Shanghai\\
  Shanghai, P.R. China\\
  \texttt{aven.le.zhou@gmail.com}\\
  \And
  Ellick Chan\\
  Northwestern University\\
  Evanston, IL, USA\\
  \texttt{ellick.chan@northwestern.edu}\\
}

\begin{document}
\maketitle

\begin{abstract}

The process of identifying and understanding art styles to discover artistic influences is essential to the study of art history. Traditionally, trained experts review fine details of the works and compare them to other known works. To automate and scale this task, we use several state-of-the-art CNN architectures to explore how a machine may help perceive and quantify art styles. This study explores: (1) How accurately can a machine classify art styles? (2) What may be the underlying relationships among different styles and artists? To help answer the first question, our best-performing model using Inception-V3 \cite{2015arXiv151200567S} achieves a 9-class classification accuracy of 88.35\%, which outperforms the model in Elgammal et al.’s study \cite{2018arXiv180107729E} by more than 20 percent. Visualizations using Grad-CAM \cite{2016arXiv161002391S} heat maps confirm that the model correctly focuses on the characteristic parts of paintings. To help address the second question, we conduct network analysis on the influences among styles and artists by extracting 512 features from the best-performing classification model. Through 2D and 3D T-SNE \cite{maaten2008visualizing} visualizations, we observe clear chronological patterns of development and separation among the art styles. The network analysis also appears to show anticipated artist level connections from an art historical perspective. This technique appears to help identify some previously unknown linkages that may shed light upon new directions for further exploration by art historians. We hope that humans and machines working in concert may bring new opportunities to the field.

\end{abstract}

% keywords can be removed
% \keywords {First keyword \and Second keyword \and More}

\section{Introduction}

\begin{large}
The process of identifying art styles and discovering artistic influences is essential to the study of art history, or more generally, the study of the visual history of humanity. Art historians have commonly sought answers to these questions through their extensive historical and cultural knowledge as well as exceptional skills in visual analysis. Now, the rise in computing power and advancement in machine learning algorithms have also allowed the machine to enter the field and contribute its own insight via data-driven methods. This study explores the potential of using convolutional neural networks (CNN) to conduct art style classification and study artistic influences among styles and artists. Several previous studies have shown the abilities of the machine in classifying art styles and generating features for analysis. However, the results of those studies fail to reflect the full capabilities of machine. Our study shows that a machine has the potential to perform art style classification better than humans without expertise in art history, and also help experts to discover underlying relationships among styles and artists.

\end{large}

\section{Methodology Overview}
\begin{large}
To determine the style of a painting, art historians may consider the depicted subject matter, brushstrokes, the use of color, perspective, and its similarities to other paintings. In the first stage of this study, we evaluate how accurately machines can classify art styles. Elgammal et al.\cite{2018arXiv180107729E} attained a 20-class classification accuracy of 63.7\% with 77K images of paintings using a pre-trained ResNet. However, we believe that a 20-class labeling may not be ideal, as there could be high similarities among certain styles, which tend to overcomplicate the problem and cause confusion for the classifier. For example, style element overlap exists between Post-Impressionism and Fauvism, as well as between Expressionism and Abstract-Expressionism. Meanwhile, we noticed that many artworks were originally mislabeled and thus require additional cleaning and polishing. To simplify the task, we reduced the number of classes to nine in order to better highlight distinctive elements between styles. We also manually relabeled some paintings to increase quality. With this curated dataset, we expect to achieve a better result even with less training data.

\par

Our dataset contains 24,110 paintings, which are mostly accurately labeled into nine classes. We use state-of-art convolutional neural networks for art style classification and trained the models on GPUs initially with a low resolution to select viable models, and retrained high performing models on CPU in higher resolution to bump up the accuracy. In particular, we implemented VGG-16 \cite{2014arXiv1409.1556S}, ResNet-152 \cite{2015arXiv151203385H}, Inception V3\cite{2015arXiv151200567S} and Inception-ResNet V2 \cite{2016arXiv160207261S} and compare their performance on the testing data. To assess whether the networks are focusing on salient parts of the images when making decisions, we also implemented Grad-CAM \cite{2016arXiv161002391S} to create heat maps. Additionally, we visualize the filters of selected convolutional layers to better understand the mechanics of the model.

% EfficientNet\cite{tan2019efficientnet}, NASNet Large\cite{nasnet},

\par
The best-performing model we found from the first stage is based on Inception V3. We use this as a proxy model in the second stage to study how machines may perceive the world. In order to explore relationships and determine similarities among styles and artists, we compute both average Euclidean and Cosine distances between every pair of artists using the output of the final fully connected layer with 512 neurons as a feature embedding for each painting. An artist network analysis is then performed to reveal Cosine linkages among nearby artists, as each artist may be considered to be connected to another artist when the distance between the two artists is minimized (for maximum similarity). Finally, we visualize the connections in chronological order to infer unidirectional influences among artists along the timeline.

\end{large}

\section{Data}
\begin{large}
We use images of paintings from the publicly available WikiArt database \cite{WEBSITE:1}, which features a total of 250,000 artworks by 3,000 artists. With a focus on nine art styles ranging from the 13th century to the 21st century, we choose 24,110 paintings from 235 artists as our dataset for this study. For the purposes of this paper, style primarily refers to the visual appearance of a painting and that does not necessarily relate it to other artists and works of any art movements. The details of our dataset are summarized in Table 1.

\par

To promote quality of the analysis, we restrict the scope of our study to a subset of data. The labeling of the styles is based on both the information on WikiArt and our best knowledge. For artists featuring works of multiple styles, we designate them to one primary style and only keep images of paintings corresponding to the selected style. We have tried our best to conduct data cleaning based on the following rules:
\par

\begin{itemize}
\item[$\ast$] Removing images that are not paintings (e.g. sculptures \& architecture)
\item[$\ast$] Removing images that contain large portions of non-paintings (e.g. frames \& walls)
\item[$\ast$] Removing images of sketches and drawings
\item[$\ast$] Removing black and white images
\item[$\ast$] Removing images with significant distortion or are of poor quality

\end{itemize}

\begin{table}[t]
  \begin{center}
    \caption{Style classes summary}
    \label{tab:table1}
    \begin{tabular}{c|c|c|c}
      \toprule % <-- Toprule here
      \textbf{Index} & \textbf{Style Class} & \textbf{\# Images} & \textbf{\# Artists Represented}\\
      \midrule % <-- Midrule here
      1 & Early Renaissance  & 1,188 & 21\\
      2 & High Renaissance & 1,442 & 25\\
      3 &   Baroque &   3,462   &  50\\
      4 &   Realism &   4,004   &   33\\
      5 &   Impressionism   &   7,788   &   20\\
      6 &   Cubism  &   1,258   &   14\\
      7 &   Abstract Art    &   2,927   &   37\\
      8 &   Pop Art &   1,050 & 24\\
      9 &   Ukiyo-e &   991   &   11\\
      \bottomrule % <-- Bottomrule here
    \end{tabular}
  \end{center}
\end{table}

It is worth noting that due to the nature of art production and conservation across the centuries, paintings of certain styles are available in greater quantities than others; thus, the classes are not quite balanced. While augmenting examples from minority classes by rotating, flipping, shifting images may be an option, we decided not to apply it so that the model can more realistically reflect the distribution of style availability in the real world.

\end{large}

\section{Art Style Classification}
\begin{large}

We randomly split 90\% of the 24,110 paintings into a training dataset and use the remaining 10\% for testing. As mentioned earlier, we experiment with four convolutional neural networks: VGG-16, ResNet-152, Inception V3, and Inception-ResNet V2. For each of the four models, the final softmax layer is replaced with a layer of nine softmax nodes, one for each style class. We train our models both with pre-trained weights from ImageNet and without pre-trained weights. It turns out that the models trained without using pre-trained weights perform better than those with pre-trained weights in terms of test accuracy, especially for ResNet-152, InceptionV3 and Inception-ResNet V2. This may be an effect of inadequate domain adaption where the domain of natural images that these models were trained on differs significantly from the abstract domain of artistic paintings.
\par

For training efficiency, we explore the tradeoff of image size on model performance, as smaller models typically train faster. After several rounds of experiments, we realize that resizing the input image smaller than 256x256 pixels did not produce accurate models. Ideally, we would like to train the models in large batches with images of high resolution. However, since we are limited to 8 GB of memory on our NVIDIA GTX 1080 GPU, we decided to use an image size of 300x300 across all models as a way to balance image size and batch size. After selecting the initial GPU-trained models, we re-trained using the Intel AI DevCloud with a Xeon Gold 6128 CPU, Intel-optimized Tensorflow, and 192 GB memory to better understand how the model may perform with higher resolution. We were able to utilize a larger 400x400 size as we were no longer limited by memory size. The model classification results are summarized in Table 2 and Table 3. The increased resolution does pay off and deliver several additional percentage points of accuracy.
\par

\begin{table}[ht]
  \begin{center}
    \caption{Performance of Four Models on Art Style Classification (GPU)}
    \label{tab:table2}
    \begin{tabular}{c|c|c|c}
      \toprule % <-- Toprule here
      \textbf{Model Name} & \textbf{Batch Size} & \textbf{Best Epoch/ Total Epochs} & \textbf{Validation Accuracy}\\
      \midrule % <-- Midrule here
      VGG-16 (300x300) &          12 & 12/20 & 0.8437\\
      ResNet-152 (300x300) &      16 & 19/25 & 0.7763\\
      Inception V3 (300x300) &    64 &   12/20   &  0.8617\\
      Inception-ResNet V2 (300x300) &   24 &   19/25   &   0.8624\\
      \bottomrule % <-- Bottomrule here
    \end{tabular}
  \end{center}
\end{table}
\begin{table}[ht]
  \begin{center}
    \caption{Performance of Four Models on Art Style Classification (CPU)}
    \label{tab:table2}
    \begin{tabular}{c|c|c|c}
      \toprule % <-- Toprule here
      \textbf{Model Name} & \textbf{Batch Size} & \textbf{Best Epoch/ Total Epochs} & \textbf{Validation Accuracy}\\
      \midrule % <-- Midrule here
      VGG-16 (400x400) & 150 & 27/38 & 0.8682\\
      Inception V3 (400x400) & 150 & 28/37 & 0.8735 \\
      Inception V3 (500x500) & 100 & 35/37 & 0.8835 \\
      Inception-ResNet V2 (400x400) &   150 &   27/36   &   0.8624 \\
      \bottomrule % <-- Bottomrule here
    \end{tabular}
  \end{center}
\end{table}

\begin{figure}[ht]
\centering
\includegraphics[width=130mm]{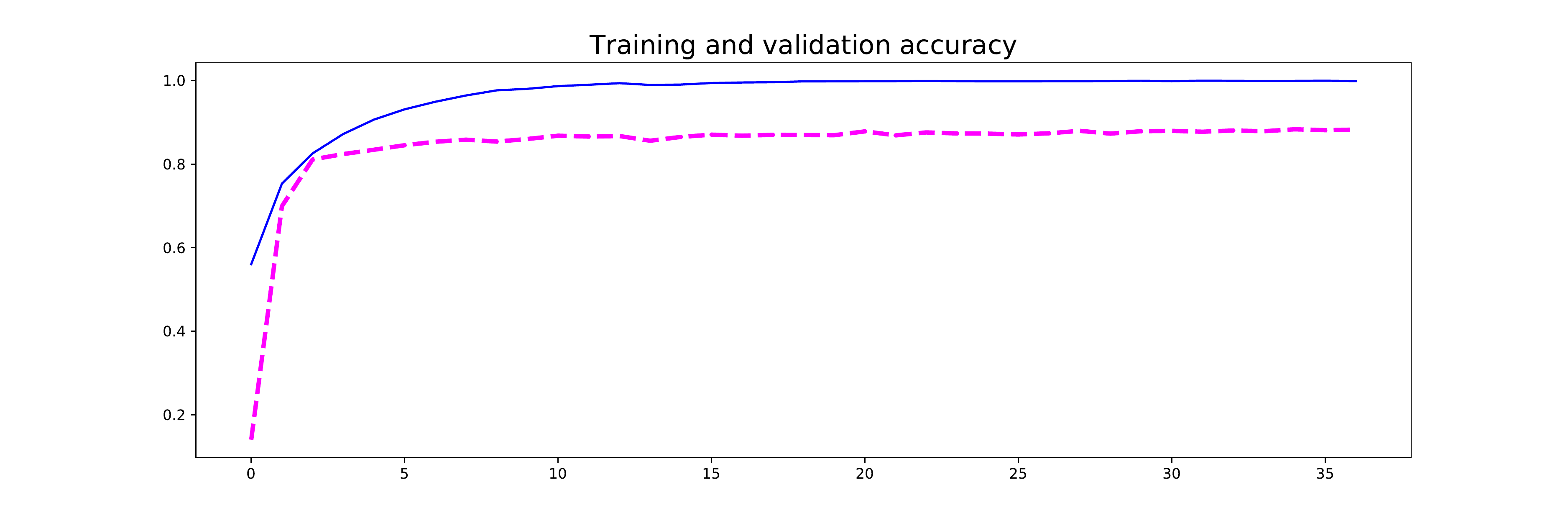}
\caption{Inception V3 Training (solid in blue) and Validation Accuracy (dashed in pink)}
\bigbreak
\includegraphics[width=130mm]{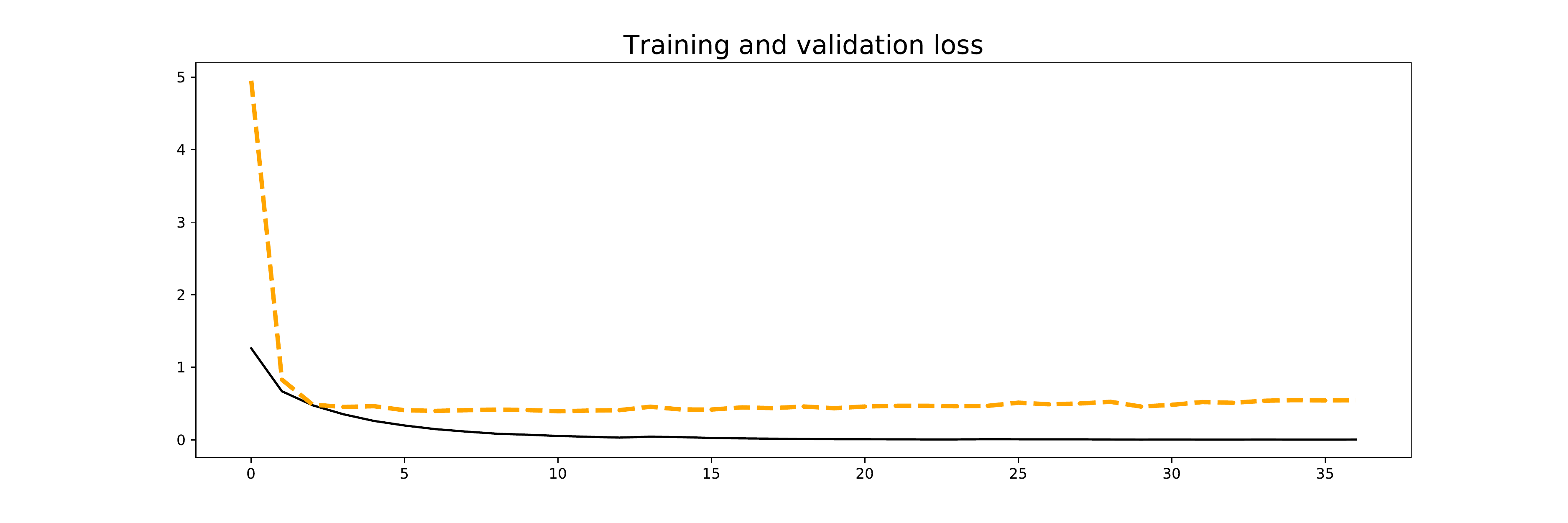}
\caption{Inception V3 Training (solid in black) and Validation Loss (dashed in yellow)}
\label{fig:figure1}
\end{figure}

As Table 3 shows, Inception V3 is the best-performing model that reaches an exciting validation accuracy of 87.35\% at an image size of 400x400, which marks a significant improvement from 63.7\% achieved by the previous study \cite{2018arXiv180107729E}. We then retrain the Inception V3 with an even larger 500x500 size and achieved an even higher validation accuracy of 88.35\%. For experts with adequate training in art history, we expect the human classification accuracy on a small sample of data would be near 100\%. However, for those with a basic understanding of art styles, we expect the human accuracy to be around 70\% to 80\%. Generally, the results exceed our prior expectations and demonstrate that a machine is able to perform the classification task better than most non-professional human beings.

\par

We hypothesize that one of the the underlying reasons behind misclassification for both human and machine is that some of the styles are quite similar aesthetically and iconically. For example, certain paintings of Early Renaissance \& High Renaissance, Baroque \& Realism, as well as Abstract Art \& Cubism share a great amount of visual similarities and might pose challenges to both human and machine alike. 

\begin{figure}[ht]
\centering
\includegraphics[width=90mm]{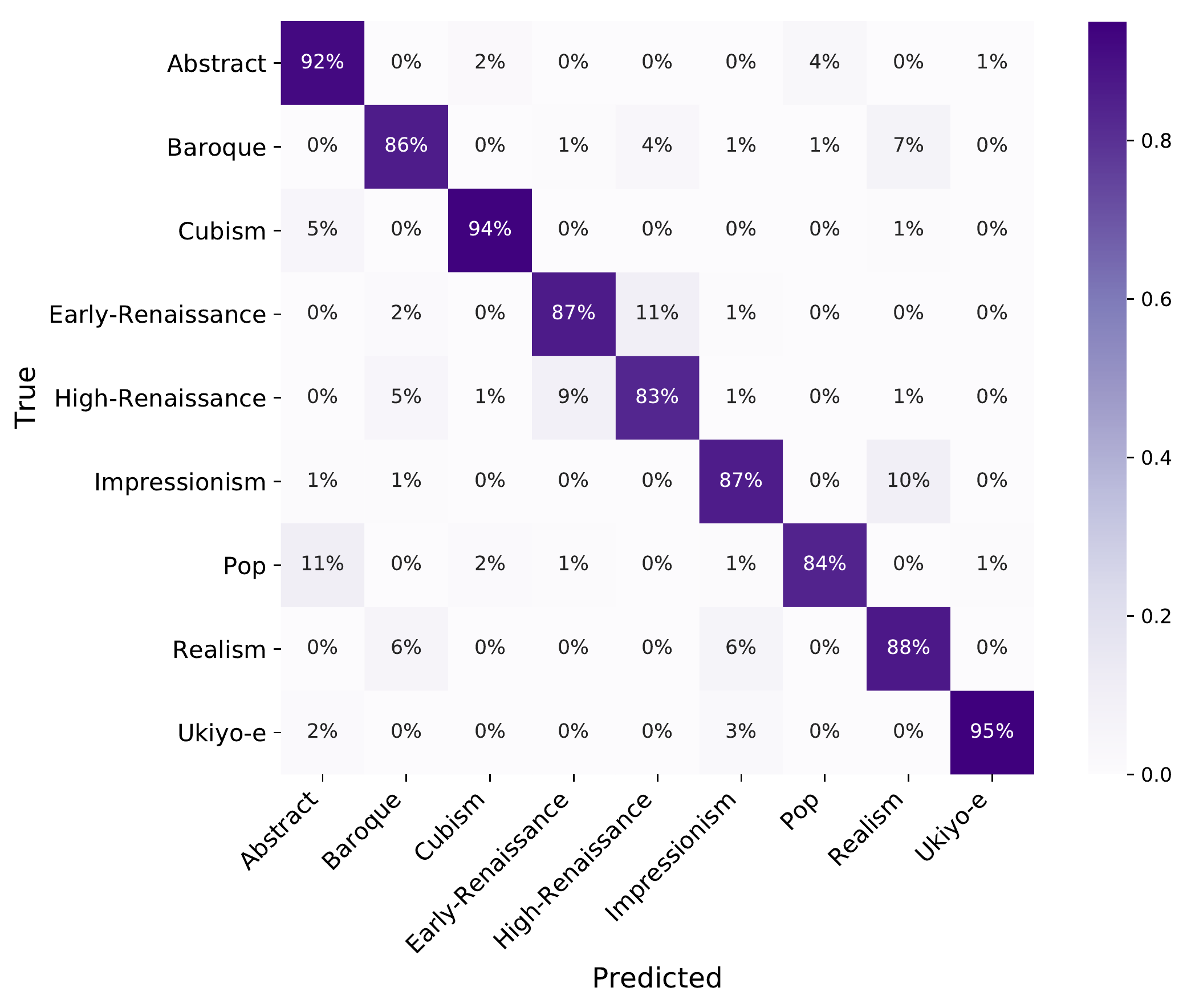}
\caption{Style Classification Confusion Matrix}
\label{fig:figure2}
\end{figure}

The confusion matrix depicted in Figure 3 largely supports this hypothesis. As it shows, the model predicts \char`\~95\% Ukiyo-e paintings correctly since this group is more aesthetically distinctive from other groups. Those styles sharing visual and iconic similarities not only pose classification challenges to humans, but to machines as well. For example, as indicated in the confusion matrix, the model predicts:

\begin{itemize}
\item[$\ast$] 26 out of 401 (\char`\~6\%) “Realism” paintings as “Baroque” paintings
\item[$\ast$] 7 out of 126 (\char`\~5\%) “Cubism” paintings as “Abstract” paintings
\item[$\ast$] 13 out of 119 (\char`\~11\%) “Early-Renaissance” paintings as “High-Renaissance” paintings
\item[$\ast$] 11 out of 105 (\char`\~11\%) “Pop” paintings as “Abstract” paintings
\end{itemize}

\begin{figure}[ht]
\centering
\captionsetup{justification=centering}
\includegraphics[clip, width=130mm]{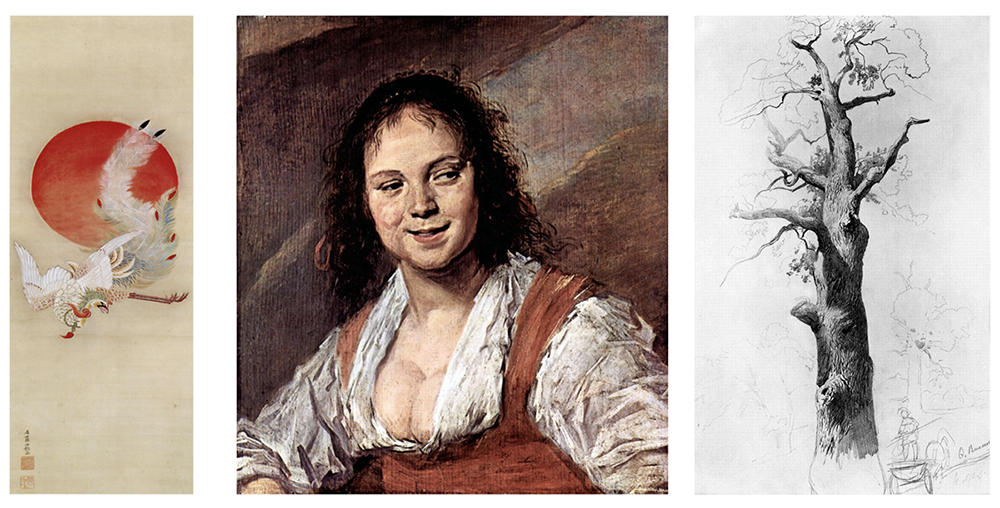}
\caption{Highlights of Misclassified Paintings: Itō Jakuchū \textit{Phoenix and Sun}, mid-Edo period (left); Frans Hal,\textit{ Portrait of a Woman}, known as \textit{The Gypsy Girl}, 1629 (middle); Fyodor Vasilyev, \textit{The Trunk of an Old Oak}, 1867-1869 (right)}
\label{fig:figure3}
\end{figure}

To delve deeper into these issues, we also manually inspected several misclassified paintings and realized that some of the misclassifications can be explained logically. For example, in Figure 4 Itō Jakuchū’s \textit{Phoenix and Sun} is predicted as abstract art with a probability of \char`\~0.52 instead of Ukiyo-e. This prediction is reasonable since the red sun and monotonic background can be interpreted as large color chunks with sparse details, which resemble elements of abstract art. Even though the two red seals on the lower left indicate the painting’s East Asian origin, additional training and tuning is necessary before the machine can recognize those specific style patterns. The second example, Frans Hal’s \textit{The Gypsy Girl} is classified as Realism with a probability of \char`\~0.87 as opposed to Baroque. We believe this decision is also reasonable as the hard brushstrokes, which seem to be made with palette knives, on the collar and sleeves are similar to the certain Realism styles. The third work, a sketch by Fyodor Vasilyev, is classified as Impressionism art with a probability of \char`\~0.58 as opposed to Realism. Here, the model actually catches a rather abstract and incomplete pencil sketch that was supposed to be removed from the dataset during data cleaning.

\par

Furthermore, to assess whether the model focuses on salient parts of images during its decision generating process, we implement Gradient-weighted Class Activation Mapping (Grad-CAM) \cite{2016arXiv161002391S} to create heat maps. Figures 5-7 highlight a few correctly-classified examples from the testing dataset.  

\par

\begin{figure}[ht]
\centering
\captionsetup{justification=centering}
\includegraphics[clip, width=130mm]{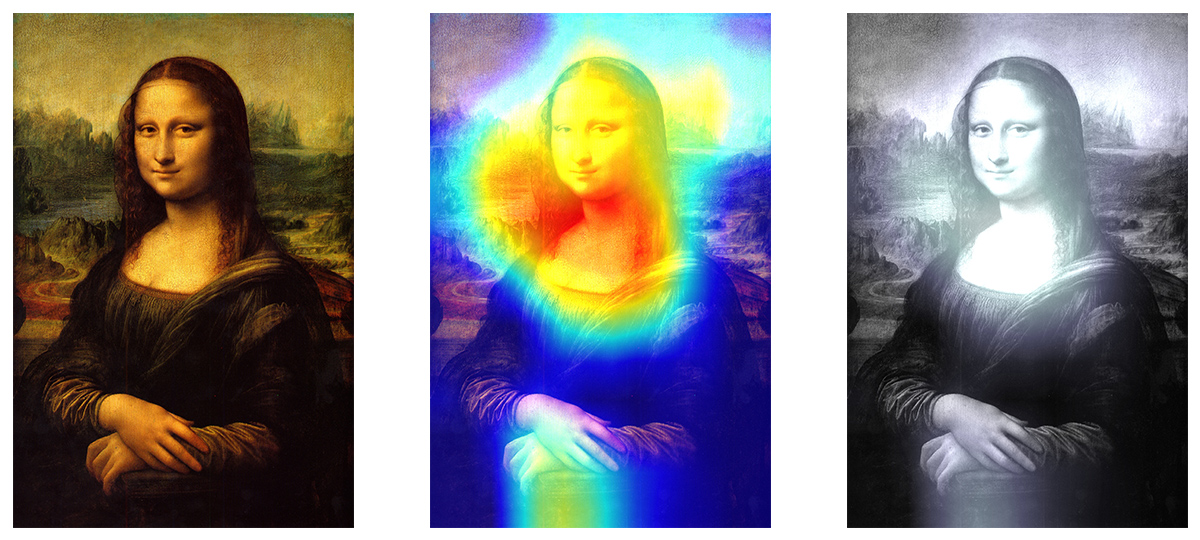}
\caption{Leonardo da Vinci, \textit{Mona Lisa}, 1503 - Correctly Predicted as High Renaissance \newline Original, Color Heat Map, and B\&W Heat Map using Grad-CAM}
% \label{fig:figure6a}
\end{figure}

\begin{figure}[ht]
\centering
\captionsetup{justification=centering}
\includegraphics[clip, width=130mm]{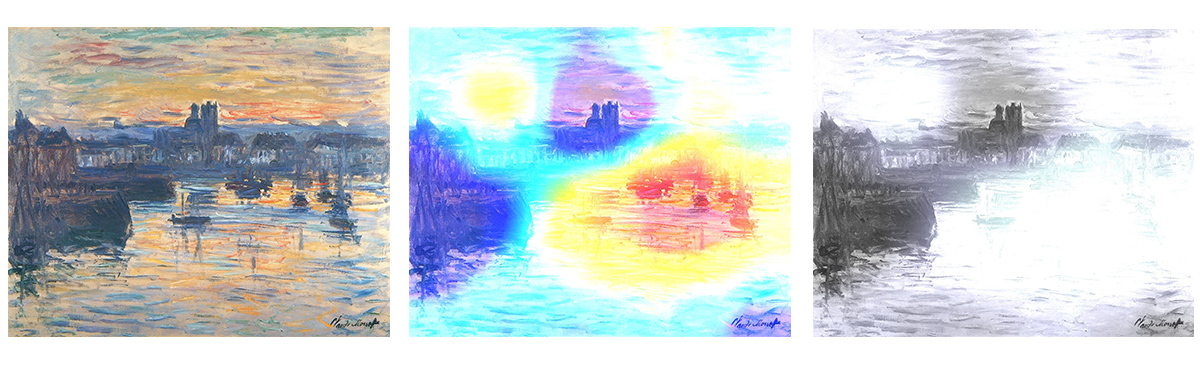}
\caption{Claude Monet, \textit{Port of Dieppe, Evening}, 1882 - Correctly Predicted as Impressionism \newline Original, Color Heat Map, and B\&W Heat Map using Grad-CAM}
% \label{fig:figure6b}
\end{figure}

\begin{figure}[ht]
\centering
\captionsetup{justification=centering}
\includegraphics[clip, width=130mm]{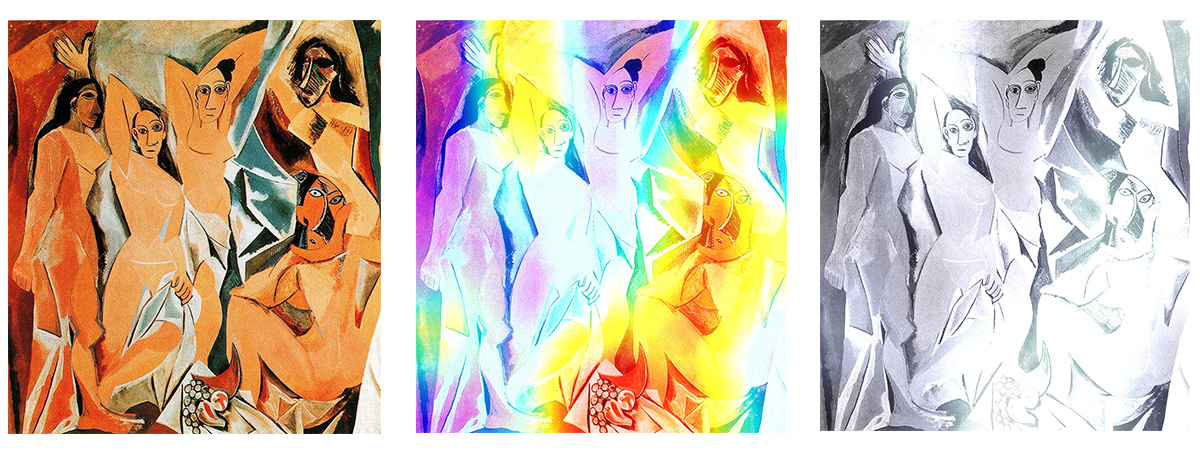}
\caption{Pablo Picasso, \textit{Les Demoiselles d'Avignon}, 1907 - Correctly Predicted as Cubism \newline Original, Color Heat Map, and B\&W Heat Map using Grad-CAM}
% \label{fig:figure6c}
\end{figure}

According to the heat maps, it seems that the model is able to focus on certain important elements. For example, in Leonardo da Vinci’s well-known \textit{Mona Lisa}, the model focuses on the woman’s face and upper chest (Figure 5). Regarding Claude Monet’s landscape \textit{Port of Dieppe, Evening}, the model pays attention to several boats on the water, wave patterns as well as the lower skyline (Figure 6). We believe the choices made here are reasonable enough as most impressionist landscape paintings do not contain a single central object. When evaluating Pablo Picasso’s \textit{Les Demoiselles d'Avignon}, the model expresses special interest in the faces and bodies of two figures on the right as well as the fruits in the foreground (Figure 7). These selected objects demonstrate clear cubic forms and mostly agree with where a human might attend to.

\par

\begin{figure}[ht]
\centering
\captionsetup{justification=centering}
\includegraphics[clip, width=130mm]{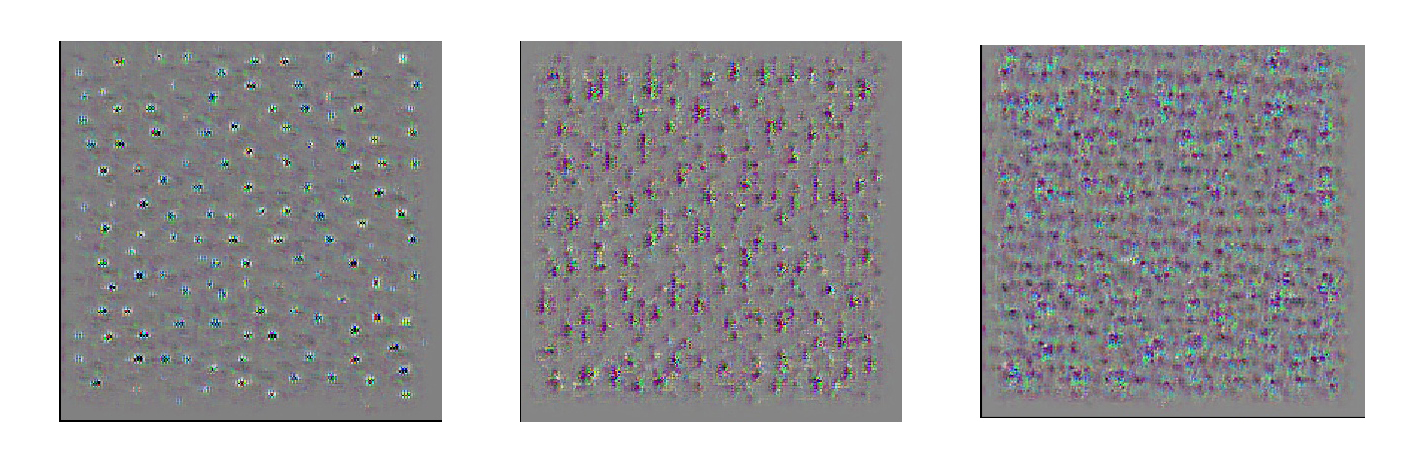}
\caption{Selected Filter Visualizations from ‘conv2d\_6’(3 out of the 48 filters)
}
% \label{fig:figure1}
\end{figure}

\begin{figure}[ht]
\centering
\captionsetup{justification=centering}
\includegraphics[clip, width=130mm]{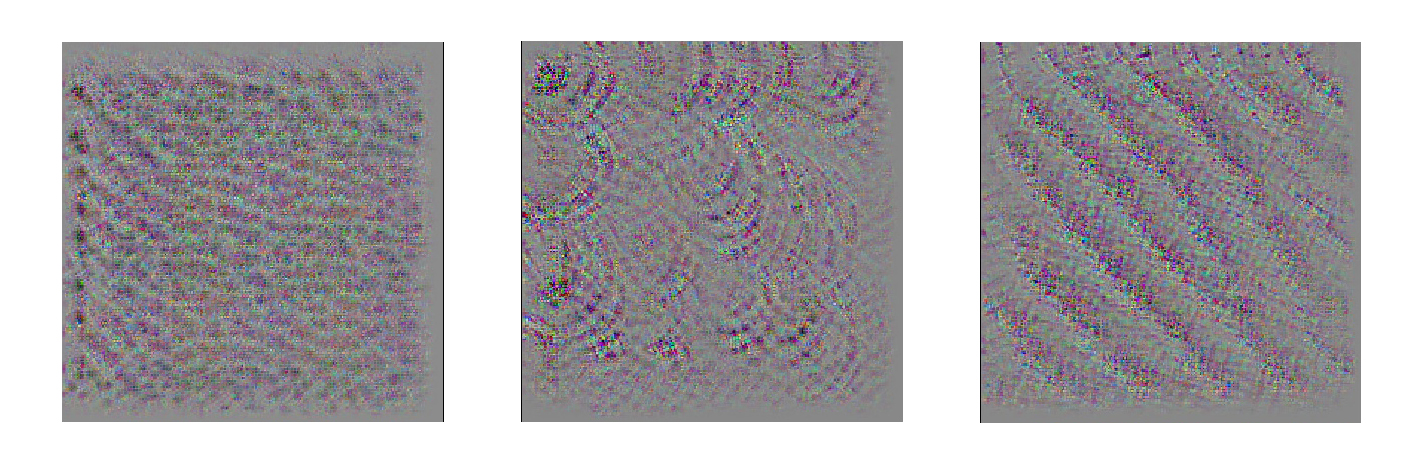}
\caption{Selected Filter Visualization from ‘conv2d\_26’ (3 out of the 384 filters)
}
% \label{fig:figure1}
\end{figure}

\begin{figure}[ht]
\centering
\captionsetup{justification=centering}
\includegraphics[clip, width=130mm]{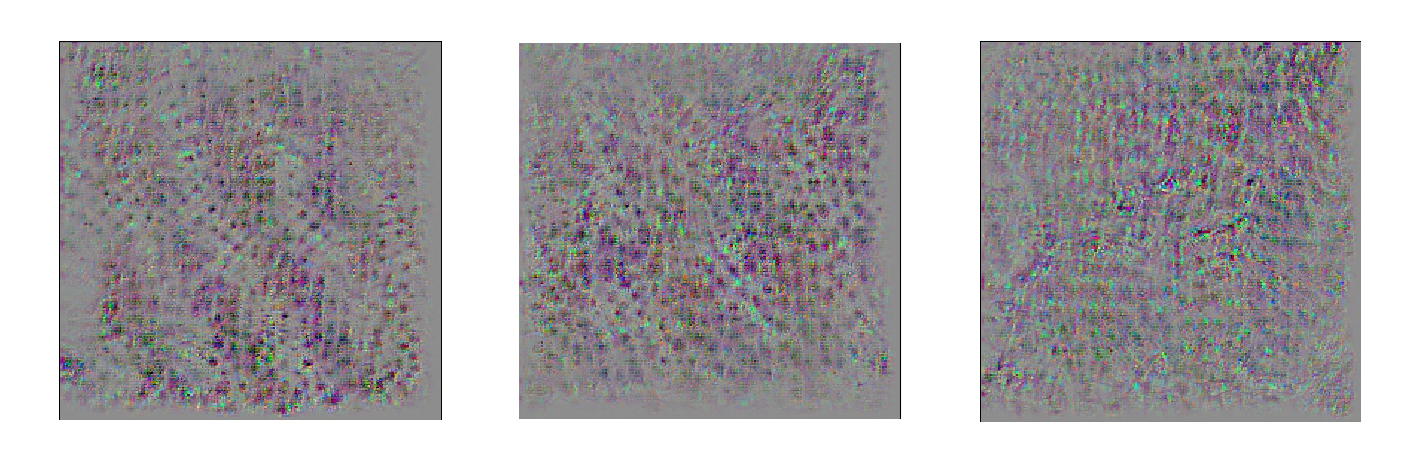}
\caption{Selected Filter Visualization from ‘conv2d\_56’ (3 out of the 160 filters)}
% \label{fig:figure1}
\end{figure}

Additionally, filter visualizations help us understand the behaviors of the network. Figure 8 shows that the network tends to focus on edges in the initial 2D convolutional layers. As the layers go deeper, the patterns on the visualizations become more abstract (Figure 9), indicating that the network can detect brushstrokes. However, visualizations of the ‘conv2d\_56’ layer (Figure 10) do not feature recognizable object patterns as we expected even after 10,000 iterations. We believe the major reasons that affects the quality of filter visualisation here are the model complexity and model saturation, which are reflected in the high training accuracy (Figure 1).
\par

\end{large}

\section{Influence Analysis - Painting Style Level}
\begin{large}

Exploring connections among painting styles and artists is the primary focus of the second stage of our study. To better understand the similarities among paintings, we first implement t-distributed Stochastic Neighbor Embedding (T-SNE) \cite{maaten2008visualizing}, a nonlinear dimensionality reduction technique well-suited for embedding high-dimensional data for visualization in a low-dimensional space. By projecting the nine-class predicted probabilities into a two-dimensional space, we are able to see that the most clusters are well separated (Figure 11).

\begin{figure}[ht]
\centering
\includegraphics[clip,trim=1cm 1cm 0cm 1cm, width=110mm]{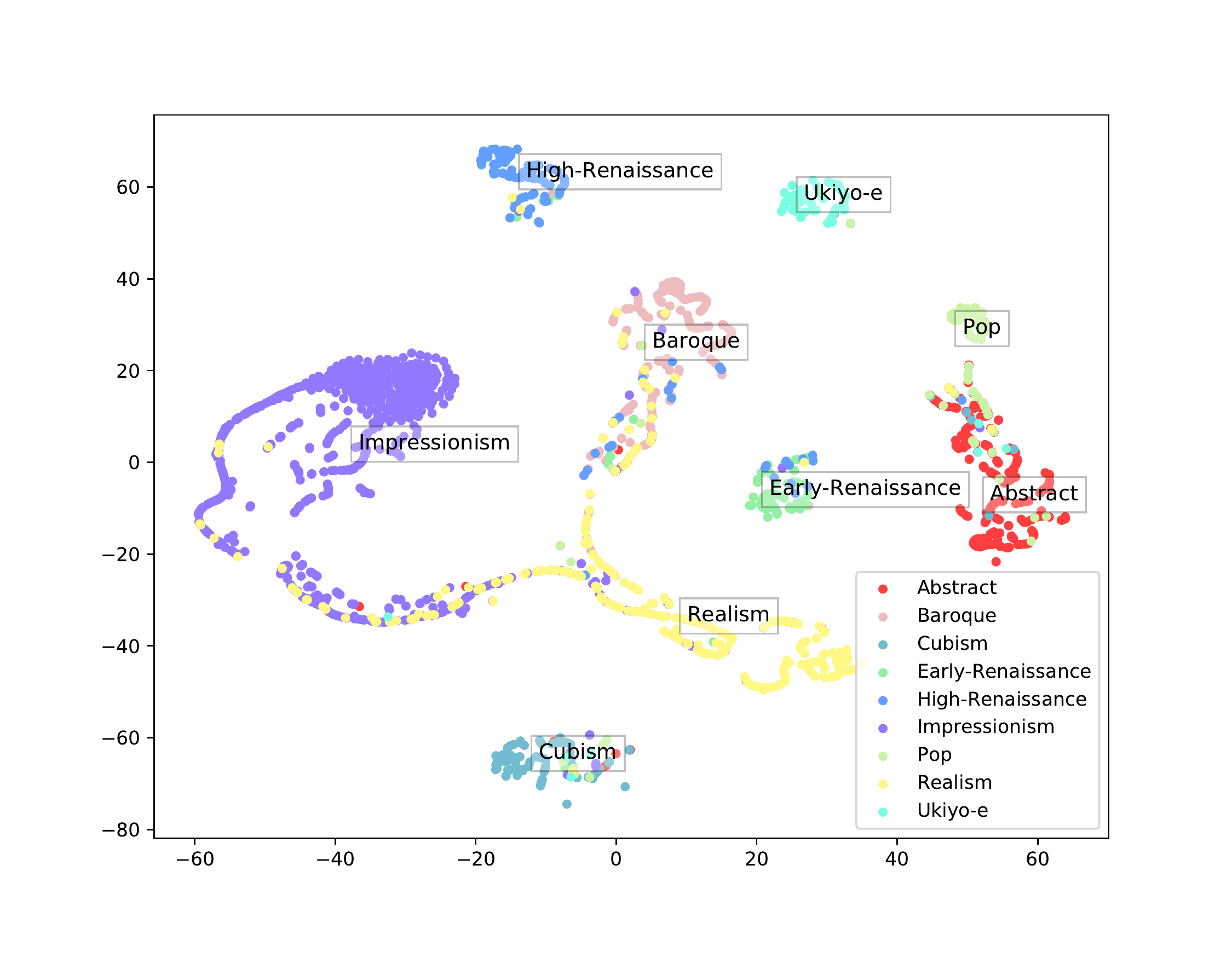}
\caption{2D T-SNE Plot on Painting Level with 9 Features}
\label{fig:figure4}
\end{figure}

\par

In order to improve model interpretability, we add a fully connected layer with 512 neurons before the softmax layers to the best-performing model based on Inception V3 in order to extract features of paintings. Aiming to expedite the re-training process and reuse as many weights from the first stage as possible, we freeze all the prior layers and only trained the fully connected layers. With the additional layer, the updated nine-class classification accuracy of the model changes to 87.77\%, which is close to highest accuracy from stage 1 (88.35\%). With this high accuracy, we believe that features extracted from the fully connected layer are representative and useful for conducting further analysis. 
\begin{figure}[ht]
\centering
\includegraphics[clip,trim=1cm 1cm 1cm 1cm, width=110mm]{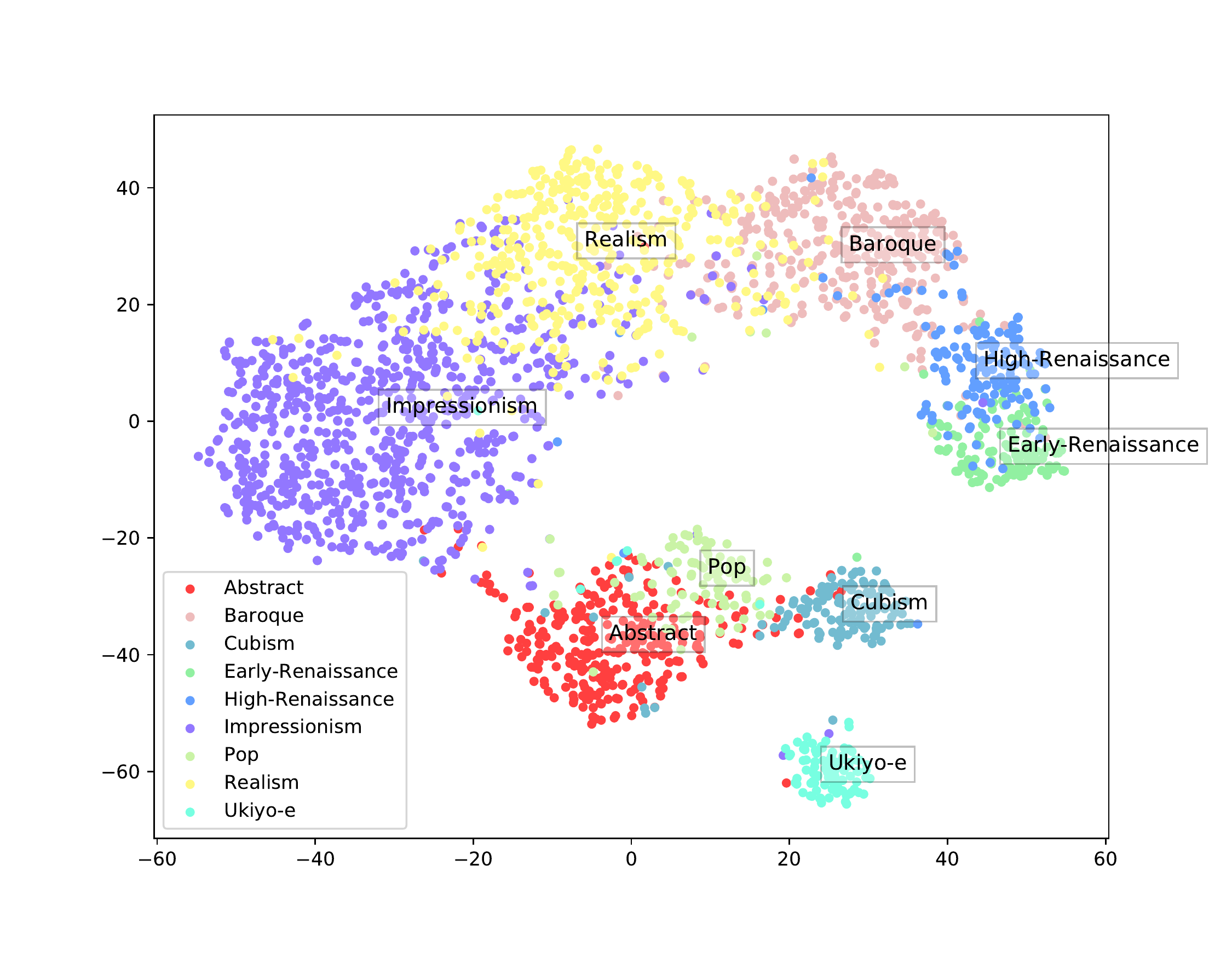}
\caption{2D T-SNE Plot on Painting Level with 512 Features}
% \label{fig:figure1}
\end{figure}

% \bigskip
\begin{figure}[ht]
\centering
\includegraphics[clip,trim=0cm 0cm 0cm 2cm, width=110mm]{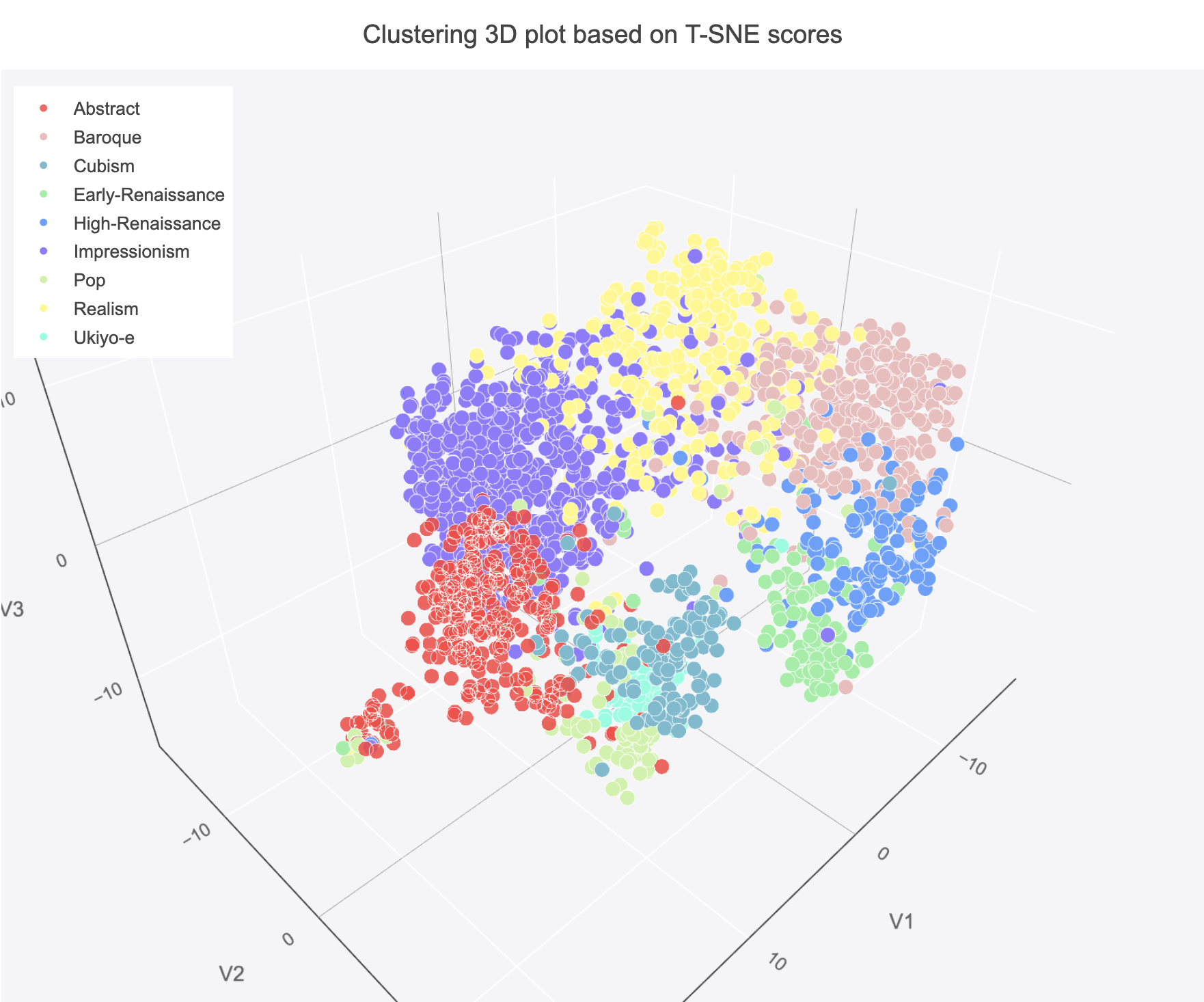}
\caption{3D T-SNE Plot on Painting Level with 512 Features}
% \label{fig:figure1}
\end{figure}

Interested in exploring connections among the 2412 paintings in the testing dataset on a painting level, we apply T-SNE to project the 512-dimensional features of each painting onto a two-dimensional space (Figure 12) and a three-dimensional space (Figure 13). The points on the 2D plot form a circular pattern around the center of the space. Focusing on the upper part of the plot, we are impressed to find that the model can uncover chronological trends of development for the five major styles. From Early-Renaissance moving in a counter-clockwise direction, High-Renaissance, Baroque, Realism and Impressionism are all sequentially following each other, depicting the developments of art styles across time. Overlapping points on the borders of the clusters may highlight style overlap during periods of transition. Additionally, clusters of Abstract art, Pop art, and Cubism are also well separated from the Impressionism cluster, indicating gaps in time and differences in style. Yet, their proximity to each other implies commonalities among those styles emerged after the 20th century. It is also worth noting that Ukiyo-e, the only non-western style category in our study, is well separated from the rest clusters, highlighting its distinctive style.

\par
\end{large}

\section{Influence Analysis - Artist Level}
\begin{large}

Additionally, we analyze influence at an artist level. To generate features for each artist, we take the average of the feature values of paintings created by each artist. In other words, each artist would have 512 features averaged across their paintings. We utilize both the Euclidean distance

\begin{figure}[p!]
\centering
\includegraphics[width=\linewidth]{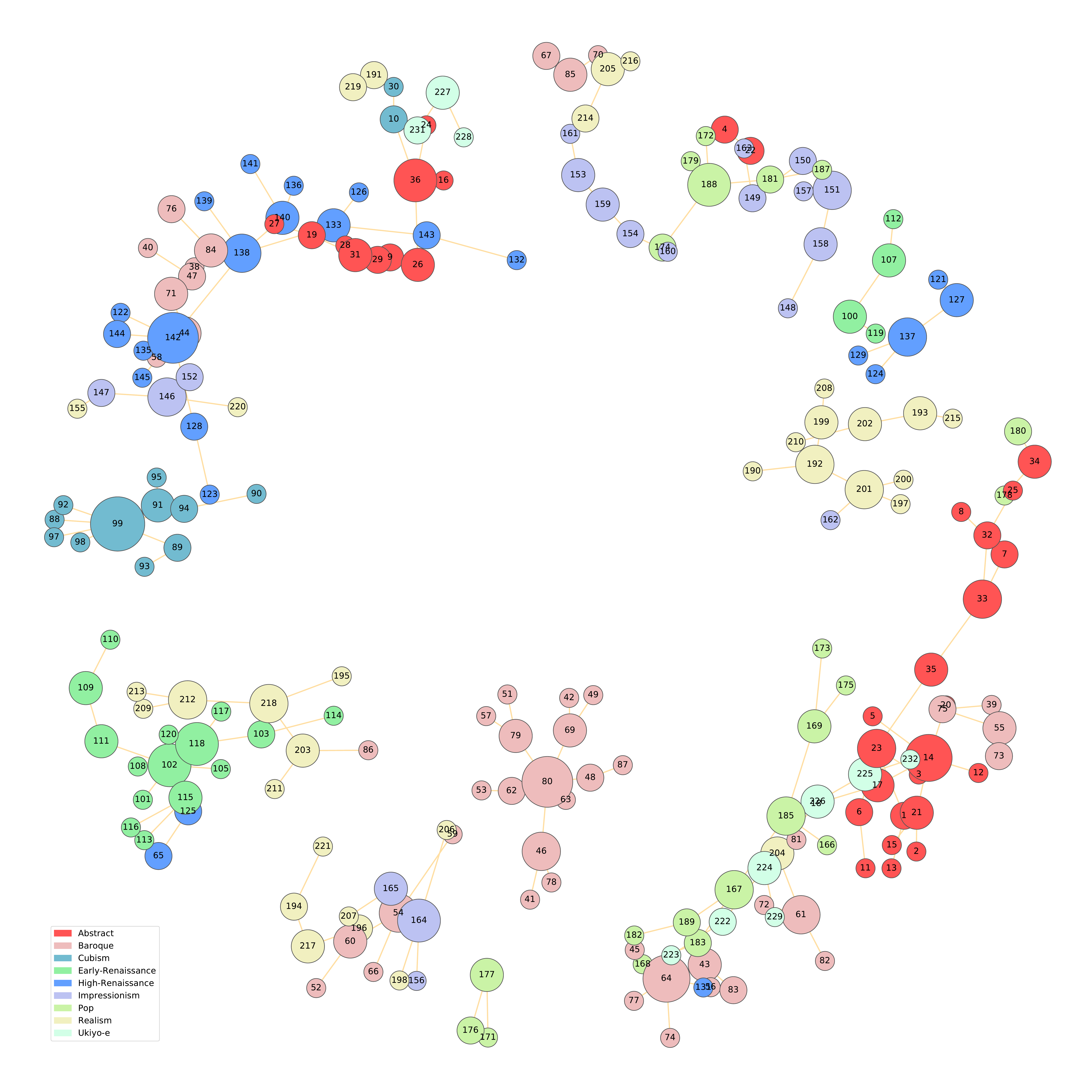}
\captionsetup{justification=centering}
\caption{Network Analysis of Artist Similarities by Style Using Cosine Distance \newline (Index-Artist Mapping Available at Appendix A)
}
% \label{fig:figure1}
\end{figure}

% \medskip
% \noindent{Euclidean Distance:}
\begin{gather*}
d(\mathbf{p},\mathbf{q}) = d(\mathbf{q}, \mathbf{p}) = \sqrt{(\mathit{q_{1}-p_{1}})^{^{2}} + (\mathit{q_{2}-p_{2}})^{^{2}} + \cdots + (\mathit{q_{n}-p_{n}})^{^{2}}  }
\end{gather*} 

\noindent and cosine similarity

% \medskip
% \noindent{Cosine Similarity:}
\begin{gather*}
\cos(\theta ) = \frac{\mathbf{A}\cdot \mathbf{B}}{\left \| \mathbf{A} \right \|\left \| \mathbf{B} \right \|} = \frac{\sum_{i=1}^{n} \mathit{A}_{i} \mathit{B}_{i}}{\sqrt{\sum_{i=1}^{n} \mathit{A}_{i}^{2}} \sqrt{\sum_{i=1}^{n}\mathit{B}_{i}^{2} }}
\end{gather*}

\noindent 
to estimate the similarity among artists. After comparing the results using two distance metrics, we realize that artist connections generated using cosine similarity are more plausible from an art historical perspective. One explanation could be that cosine similarity is more suitable for handling high-dimensional spaces, which enables it to better capture the similarities among artists.
\par

Next, we perform network analysis to visualize the global connections and influences among artists. An artist A is considered connected to artist B if this pair has the maximum cosine similarity among all possible pairs involving A. Such connections are represented as linkages in the network graph (Figure 14). While each node on the graph represents an artist, the size of a particular node is proportional to the number of edges connecting with it. Different colors of the nodes correspond to different style classes that artists belong to. 

\par

Several linkages generated by the network are consistent with art historical explanations. From the network graph, we notice that Pablo Picasso (Index 99) is closely connected with Georges Braque (Index 92). This is reasonable since they worked closely in the early 20th century and developed Cubism together. To some extent, the style of their Cubist works were indistinguishable for years. As another example, the linkages among Duccio (Index 107), Ambrogio Lorenzetti (Index 100) and Pietro Lorenzetti (Index 119) also accord with art historical facts. Duccio and Pietro Lorenzetti is connected through Ambrogio Lorenzetti. While Pietro Lorenzetti was the younger brother of Ambrogio Lorenzetti, he was also a follower of Duccio and many believe that he also studied under Duccio. Additionally, the network also correctly discovers the connections among the French Impressionists. For example, Pierre-Auguste Renoir (Index 163) is connected with Camille Pissarro (Index 150) though Berthe Morisot (Index 149), and Mary Cassatt (Index 159) is connected to Edouard Manet (Index 154) and Edgar Degas (Index 153). This body of artists played crucial roles in the Impressionism movement and held series of independent exhibitions in the second half of the 19th century. Regarding abstract art, Paul Klee (Index 26) and Wassily Kandinsky (Index 36), who are regarded as the founding fathers of classical modernism and abstract art, are also connected on the graph. From \textit{Der Blaue Reiter} avant-garde movement to Bauhaus, the two close friends critically engaged with each other’s work and explore new possibilities of virtual means. 
\par

Furthermore, certain non-obvious linkages that cannot be explained by existing studies in art history may shed light upon new directions for further exploration. For example, William Dobson (Index 87), a portraitist and one of the first notable English painters appears to be closely related to Caravaggio (Index 48), the famous Baroque artist known for his dramatic use of chiaroscuro. However, few works in the literature have explored the influence of Caravaggio on William Dobson; thus, this may be an interesting linkage that people may have overlooked. In addition, Théodore Rousseau (Index 217), a French landscape painter of the Barbizon school might have influenced Fyodor Vasilyev (Index 196), a Russian lyrical landscape painter. The hue, style, and subject matters of their paintings share a significant amount of visual similarities; yet, few people have studied possible connections between them yet.

\begin{figure}[ht]
\centering
\includegraphics[clip,trim=30cm 14cm 30cm 18cm, width=\linewidth]{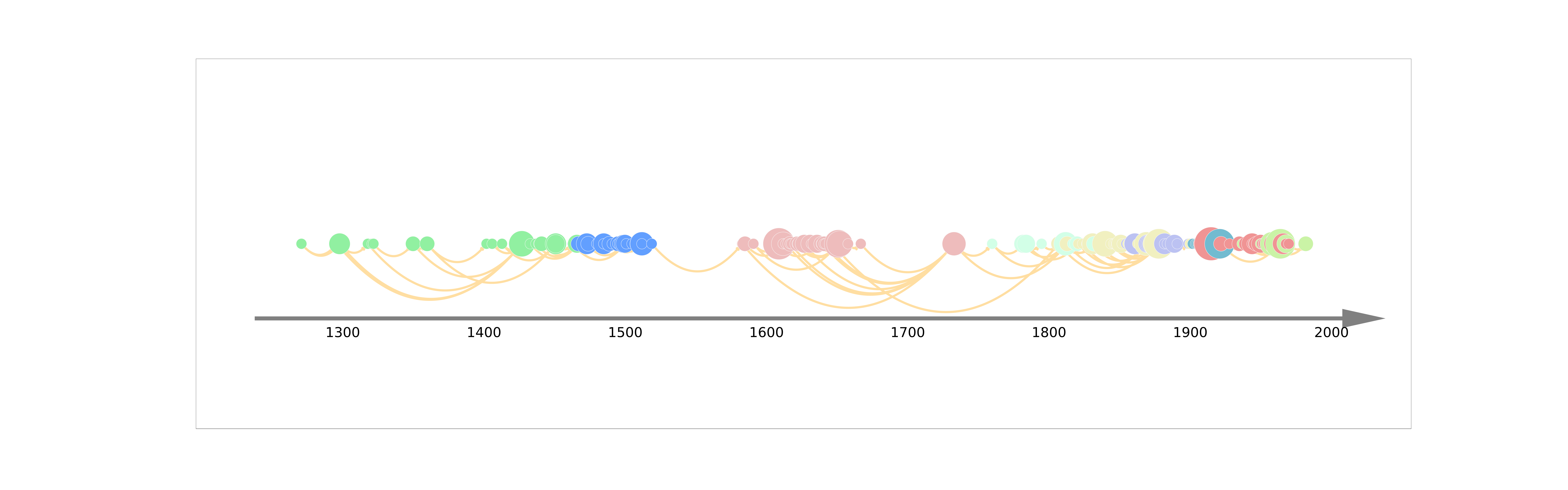}
\caption{Network Analysis of Artist Similarities by Styles Using Cosine Distance
}
% \label{fig:figure1}
\end{figure}

\par
Finally, in order to better uncover the influences among artists with respect to time, we also visualize the chronological artistic lineage by plotting the network graph along the timeline (Figure 15). The year corresponding with each artist is calculated by averaging production years of all her or his works in the dataset. The connections here were restricted to be unidirectional, since naturally only earlier artists can influence later ones. Certain generated linkages are also plausible from an art historical perspective. For example, Annibale Carracci is not only connected with his brother Agostino Carracci, but also with Caravaggio in Figure 16(a). Even though he and Caravaggio were rivals, they both reacted against the Mannersism and and favored a more naturalistic style. Additionally, while Giotto was a student of Cimabue, Duccio was also a follower of Cimabue and many believe that he also studied under Cimabue Figure 16(b). The three important artists all played important roles in the rise of individualism during the Early-Renaissance period.

\begin{figure}[htb]
\centering
\subfloat[Around 1600s, Baroque]{{\includegraphics[width=70mm]{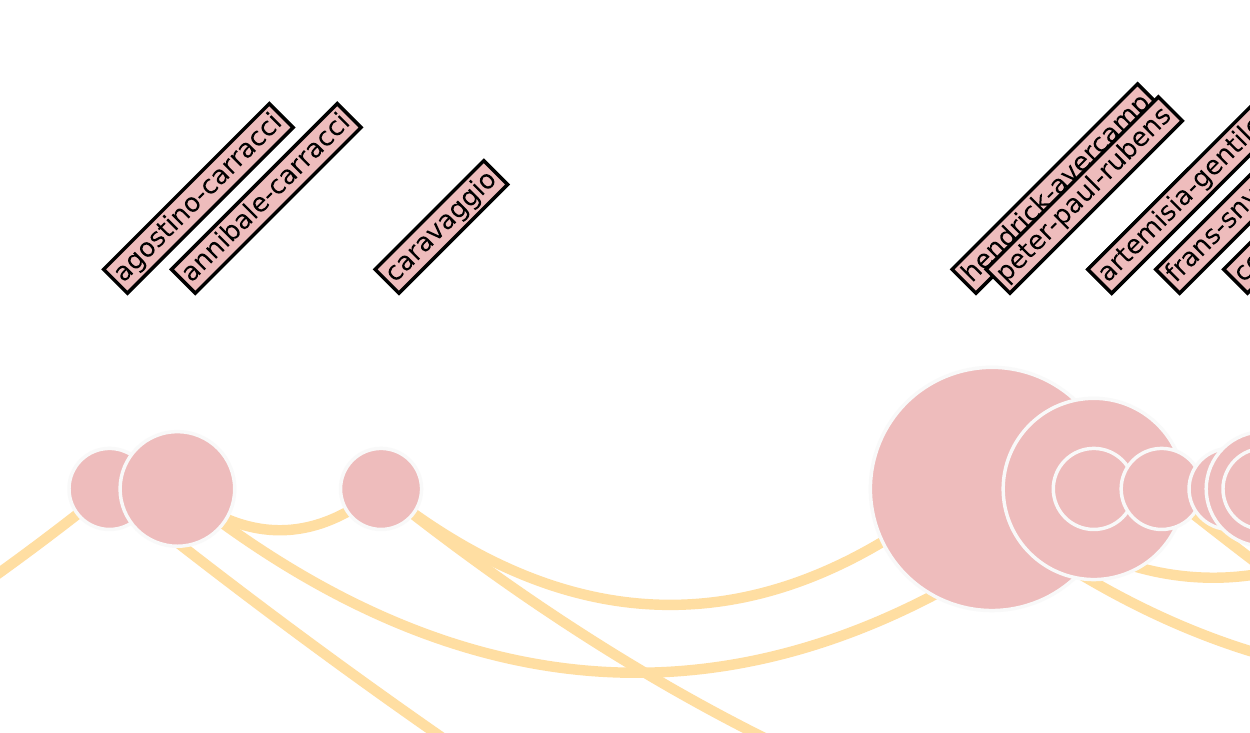}}}
\qquad
\subfloat[Around 1300s, Early Renaissance]{{\includegraphics[width=70mm]{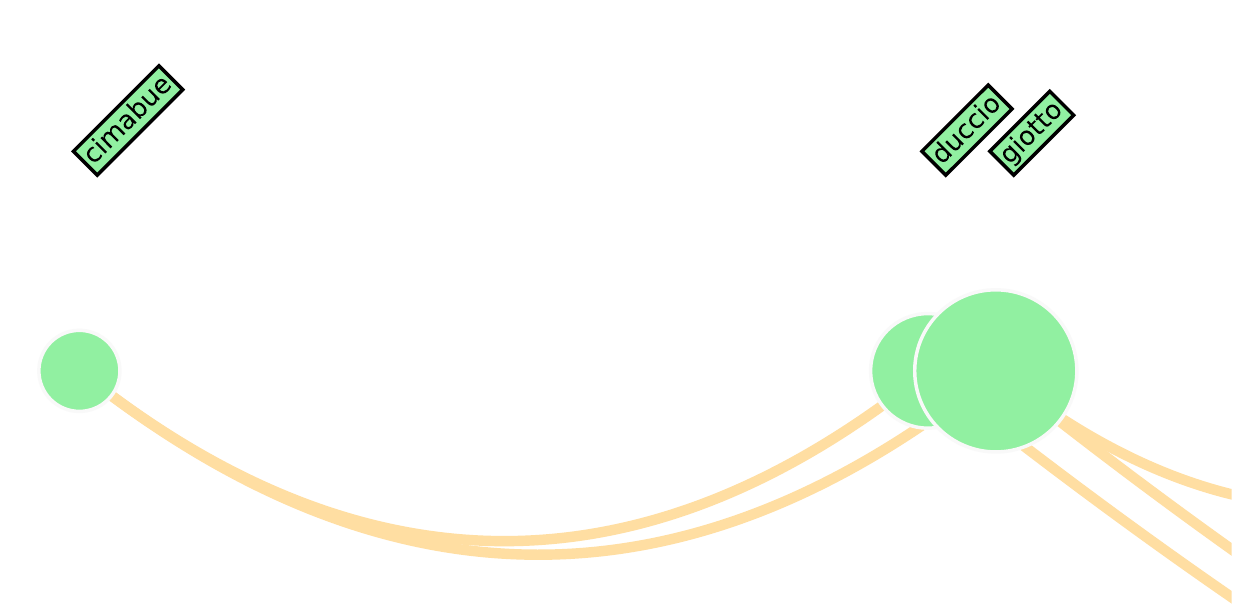}}}
\caption{Detail, Network Analysis of Artist Similarities by Styles Using Cosine Distance}
% \label{fig:figure1}
\end{figure}

\begin{figure}[thb]
\centering
\includegraphics[width=100mm]{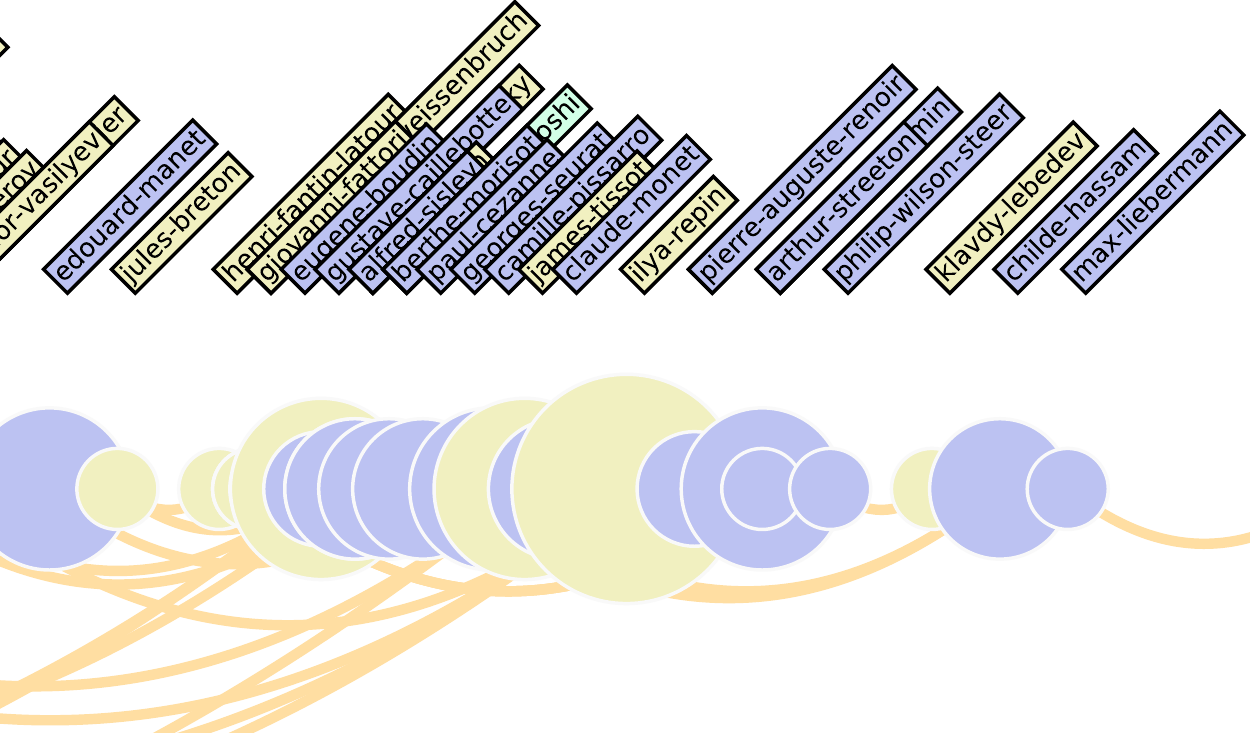}
\caption{1850 - 1900, Primarily Realism and Impressionism}
% \label{fig:figure1}
\end{figure}

\par
However, limitations do exist in our current approach. First, using the average of available production years may not be ideal especially when analyzing artists with their close contemporaries. In Figure 17, both Monet and Renoir were leading painters in the development of Impressionist style. Yet, since a high portion of works by Renoir in the dataset were created relatively later than Monet’s works, Renoir inaccurately appears to be a follower of Monet according to the plot. Secondly, for each artist A, we only pair it with one other artist, if their pairwise similarity is the highest among all possible pairs involving A. Thus, the results may not be able to capture the actual degree of complex interactions among artists. Nevertheless, the study does show the potential of using computational means to discover influences among artists.

\end{large}

\section{Conclusion}

\begin{large}
In this study, we explore the potential of using convolutional neural networks (CNN) to semi-automatically classify paintings by styles and discover relationships among styles and artists. The model based on Inception V3, our best-performing algorithm reaches an exciting validation accuracy of 88.35\%. Heat maps created using Grad-CAM confirm that the model successfully focuses on reasonable painting components during its decision-making process.

\par

Moreover, we extract features from our classification model and perform network analysis looking for patterns of art style developments. From 2D and 3D T-SNE visualizations, we discover both chronological trends of development as well as distinctiveness among styles. On an artist level, our graphs based on the cosine similarity metric reveal not only certain connections consistent with the extensive studies of art history, but also non-obvious ones, which may point to new directions for further exploration by art historians. Even though there are still opportunities for further enhancements, our study has shown the great potential of using deep learning to help experts derive data-driven insights for art historical questions.

% \end{large}
\bibliographystyle{unsrt}

\end{large}
\newpage 
\appendix
\section{Index - Artist Mapping}

% \begin{tabular}{l|c}%
%     \bfseries Index & \bfseries Artist
%     \csvreader[head to column names]{artist_mapping.csv}{}
%     {\\\hline\Index & \Artist}
%     \end{tabular}

\begin{multicols}{2}

\setbox\ltmcbox\vbox{
\makeatletter\col@number\@ne

\begin{longtable}{c|c}%
    \bfseries Index & \bfseries Artist
    \begin{filecontents*}{artist_mapping.csv}
Index,Artist
1,Adolph Gottlieb
2,Agnes Martin
3,Arshile Gorky
4,Barnett Newman
5,Brice Marden
6,Clyfford Still
7,Cy Twombly
8,Elaine De Kooning
9,Esteban Vicente
10,Francis Picabia
11,Franz Kline
12,Friedel Dzubas
13,Hans Hofmann
14,Helen Frankenthaler
15,Jack Bush
16,Jackson Pollock
17,James Brooks
18,Jean Paul Riopelle
19,Joan Mitchell
20,John Mclaughlin
21,Kenneth Noland
22,Mark Rothko
23,Mark Tobey
24,Norman Bluhm
25,Paul Jenkins
26,Paul Klee
27,Philip Guston
28,Richard Diebenkorn
29,Richard Pousette Dart
30,Robert Delaunay
31,Robert Goodnough
32,Robert Motherwell
33,Ronnie Landfield
34,Sonia Delaunay
35,Theodoros Stamos
36,Wassily Kandinsky
37,William Turnbull
38,Adriaen Brouwer
39,Adriaen Van Ostade
40,Aelbert Cuyp
41,Agostino Carracci
42,Alonzo Cano
43,Annibale Carracci
44,Anthony Van Dyck
45,Artemisia Gentileschi
46,Bartolome Esteban Murillo
47,Bernardo Strozzi
48,Caravaggio
49,Charles Le Brun
50,Clara Peeters
51,Claude Lorrain
52,Claudio Coello
53,Cornelis De Vos
54,David Teniers The Younger
55,Diego Velazquez
56,Eustache Le Sueur
57,Frans Francken The Younger
58,Frans Hals
59,Frans Snyders
60,Gabriel Metsu
61,Gerard Terborch
62,Gerrit Dou
63,Giovanni Battista Tiepolo
64,Guido Reni
65,Hans Holbein The Younger
66,Hendrick Avercamp
67,Hercules Seghers
68,Hyacinthe Rigaud
69,Jacob Jordaens
70,Jacob Van Ruisdael
71,Jan Steen
72,Jan Van Goyen
73,Johannes Vermeer
74,Juan Carreno De Miranda
75,Judith Leyster
76,Jusepe De Ribera
77,Le Nain Brothers
78,Mattia Preti
79,Nicolas Poussin
80,Peter Paul Rubens
81,Pieter De Hooch
82,Pieter Van Hanselaere
83,Pietro Da Cortona
84,Rembrandt
85,Salvator Rosa
86,Willem Kalf
87,William Dobson
88,Albert Gleizes
89,Amadeo De Souza Cardoso
90,Aristarkh Lentulov
91,Fernand Leger
92,Georges Braque
93,Georges Valmier
94,Gino Severini
95,Gosta Adrian Nilsson
96,Henri Le Fauconnier
97,Jean Metzinger
98,Lyubov Popova
99,Pablo Picasso
100,Ambrogio Lorenzetti
101,Andrea Del Castagno
102,Benozzo Gozzoli
103,Carlo Crivelli
104,Cennino Cennini
105,Cimabue
106,Domenico Veneziano
107,Duccio
108,Filippo Lippi
109,Fra Angelico
110,Gentile Da Fabriano
111,Giotto
112,Giovanni Da Milano
113,Lo Scheggia
114,Lorenzo Monaco
115,Luca Di Tomme
116,Masaccio
117,Paolo Uccello
118,Piero Della Francesca
119,Pietro Lorenzetti
120,Simone Martini
121,Andrea Del Sarto
122,Andrea Del Verrocchio
123,Andrea Mantegna
124,Andrea Solario
125,Bartolomeo Veneto
126,Cima Da Conegliano
127,Correggio
128,Domenico Ghirlandaio
129,Dosso Dossi
130,Fra Bartolomeo
131,Giorgione
132,Giovanni Antonio Boltraffio
133,Giovanni Bellini
134,Giulio Romano
135,Il Sodoma
136,Leonardo Da Vinci
137,Lorenzo Lotto
138,Luca Signorelli
139,Mariotto Albertinelli
140,Michelangelo
141,Piero Di Cosimo
142,Pietro Perugino
143,Pinturicchio
144,Raphael
145,Sandro Botticelli
146,Alfred Sisley
147,Armand Guillaumin
148,Arthur Streeton
149,Berthe Morisot
150,Camille Pissarro
151,Childe Hassam
152,Claude Monet
153,Edgar Degas
154,Edouard Manet
155,Eugene Boudin
156,Frederic Bazille
157,Georges Seurat
158,Gustave Caillebotte
159,Mary Cassatt
160,Max Liebermann
161,Paul Cezanne
162,Philip Wilson Steer
163,Pierre Auguste Renoir
164,Tom Roberts
165,Walter Sickert
166,Allan D Arcangelo
167,Andy Warhol
168,Antonio Berni
169,Billy Apple
170,Bruno Munari
171,Derek Boshier
172,Eduardo Paolozzi
173,Edward Ruscha
174,Erro
175,Evelyne Axell
176,Jasper Johns
177,Larry Rivers
178,Mario Schifano
179,Marjorie Strider
180,Martial Raysse
181,Patrick Caulfield
182,Pauline Boty
183,Peter Blake
184,Peter Phillips
185,Richard Hamilton
186,Robert Rauschenberg
187,Rosalyn Drexler
188,Roy Lichtenstein
189,Tom Wesselmann
190,Adolph Menzel
191,Alexey Venetsianov
192,Camille Corot
193,Charles Francois Daubigny
194,Constant Troyon
195,Ernest Meissonier
196,Fyodor Vasilyev
197,George Catlin
198,Giovanni Fattori
199,Gustave Courbet
200,Henri Fantin Latour
201,Ilya Repin
202,Ivan Shishkin
203,James Tissot
204,Jean Baptiste Simeon Chardin
205,Jean Francois Millet
206,Johan Hendrik Weissenbruch
207,John French Sloan
208,Jules Bastien Lepage
209,Jules Breton
210,Jules Dupre
211,Klavdy Lebedev
212,Konstantin Makovsky
213,Nikolai Ge
214,Pavel Fedotov
215,Rosa Bonheur
216,Theodore Gericault
217,Theodore Rousseau
218,Vasily Perov
219,Vasily Tropinin
220,Wilhelm Leibl
221,William Simpson
222,Hiroshige
223,Ito Jakuchu
224,Katsushika Hokusai
225,Keisai Eisen
226,Kitagawa Utamaro
227,Toshusai Sharaku
228,Toyota Hokkei
229,Tsukioka Yoshitoshi
230,Utagawa Kunisada Ii
231,Utagawa Kunisada
232,Utagawa Toyokuni
\end{filecontents*}
\csvreader[head to column names]{artist_mapping.csv}{}
    {\\\hline\Index & \Artist}
    
    \end{longtable}
\unskip
\unpenalty
\unpenalty
}
\unvbox\ltmcbox

\end{multicols}

\end{document}